\documentclass[runningheads]{llncs}
\usepackage[T1]{fontenc}
\usepackage{paralist}
\usepackage{hyperref}
\usepackage{algorithm}
\usepackage{algorithm}
\usepackage{balance}
\usepackage{amsmath}
\usepackage{amssymb}
\usepackage{multirow}
\usepackage{mathtools}
\usepackage{algpseudocode}
\usepackage{textcomp}
\usepackage{gensymb}
\usepackage{algorithm}
\usepackage{graphicx}
\usepackage{subcaption}
\usepackage{mwe}
\usepackage[normalem]{ulem}
\usepackage{color}
\usepackage{soul}
\usepackage{ragged2e}
\usepackage{multirow}
\usepackage{subcaption}
\usepackage[table]{xcolor}
\usepackage{colortbl}

\usepackage{lipsum}

\usepackage{forloop}
\newcounter{ct}

\definecolor{darkblue}{RGB}{0,0,120}
\definecolor{urlblue}{RGB}{32,43,255}
\definecolor{persoRed}{RGB}{175,35,24}
\usepackage{paralist}

\begin{document}
%
% \title{Interpretable Counterfactual Explanations for Time Series Data Guided by Shapelets}
\title{Info-CELS: Informative Saliency Map Guided Counterfactual Explanation for Time Series Classification}

\titlerunning{Info-CELS: Informative Saliency Map Guided Counterfactual Explanation}
\author{Peiyu Li\inst{1} \and
Omar Bahri\inst{1} \and
Pouya Hosseinzadeh\inst{1} \and
Souka\"ina Filali Boubrahimi\inst{1} \and
Shah Muhammad Hamdi\inst{1}}
\authorrunning{P. Li et al.}
\institute{Department of Computer Science, 
Utah State University,
 Logan, UT 84322, USA 
 Emails: \texttt{\{peiyu.li, omar.bahri, pouya.hosseinzadeh, soukaina.boubrahimi,  s.hamdi\}@usu.edu}
% \email{peiyu.li@usu.edu}\\
% \email{}
}
\maketitle              % typeset the header of the contribution
\begin{abstract}
As the demand for interpretable machine learning approaches continues to grow, there is an increasing necessity for human involvement in providing informative explanations for model decisions. This is necessary for building trust and transparency in AI-based systems, leading to the emergence of the Explainable Artificial Intelligence (XAI) field. Recently, a novel counterfactual explanation model, CELS, has been introduced. CELS learns a saliency map for the interest of an instance and generates a counterfactual explanation guided by the learned saliency map. While CELS represents the first attempt to exploit learned saliency maps not only to provide intuitive explanations for the reason behind the decision made by the time series classifier but also to explore post hoc counterfactual explanations, it exhibits limitations in terms of high validity for the sake of ensuring high proximity and sparsity. In this paper, we present an enhanced approach that builds upon CELS. While the original model achieved promising results in terms of sparsity and proximity, it faced limitations in validity. Our proposed method addresses this limitation by removing mask normalization to provide more informative and valid counterfactual explanations. Through extensive experimentation on datasets from various domains, we demonstrate that our approach outperforms the CELS model, achieving higher validity and producing more informative explanations.
\end{abstract}

\section{Introduction}
\label{sec: intro}
In recent years, time series classification methods have seen extensive adoption across various fields \cite{chaovalitwongse2006electroencephalogram}, \cite{hosseinzadeh2023metforc}, \cite{bahri2022shapelet}, \cite{li2022fast}. However, many of these methods, especially those employing deep learning techniques \cite{ismail2019deep}, prioritize accuracy over interpretability. This focus on accuracy without transparent decision-making mechanisms presents challenges, particularly in domains where explanations are vital for trust. Understandably, domain experts are hesitant to rely solely on machine-driven decisions and seek insights into the rationale behind recommendations \cite{adadi2018peeking}. Establishing trust between humans and decision-making systems is crucial, and achievable through interpretable explanations during model development or via post-hoc explanation methods, enhancing transparency and building trust in decision-making processes, especially in domains where human judgment is paramount.

To address the transparency issue in machine learning models, experts advocate for Explainable Artificial Intelligence (XAI) methods \cite{tjoa2020survey}. XAI aims to provide reliable explanations for model decisions while maintaining high performance.

One prominent post-hoc explanation approach is to determine feature attributions given a prediction through local approximation. Ribeiro et al. \cite{ribeiro2016should} propose LIME (Local Interpretable Model-Agnostic Explanations), which examines the effect of perturbing the input near the decision boundary to generate insights. Guidotti et al. \cite{guidotti2018local} extend LIME with LORE (Local Rule-based Explanations), a model-agnostic explanation approach based on logic rules. Lundberg and Lee \cite{lundberg2017unified} present SHAP (SHapley Additive exPlanations), assigning each feature an importance value based on cooperative game theory's Shapley values.

Visualizing the model's decision is another common technique for explaining predictions \cite{mothilal2020explaining}. In computer vision, visualization techniques like highlighting important image parts or generating class activation maps (CAM) in convolutional neural networks \cite{mahendran2015understanding} provide intuitive explanations.

Recently, counterfactual (CF) explanation methods have gained attention as an alternative to feature-based approaches. The first CF explanation model, wCF, was introduced by \cite{wachter2017counterfactual} in the context of XAI. wCF minimizes a loss function comprising two components: a prediction loss term that drives the counterfactual to change the model's outcome, and a distance loss term that ensures the counterfactual remains close to the original input using Manhattan distance. Building on wCF, several optimization-based algorithms, such as SG \cite{li2022sg} and TeRCE \cite{bahri2022temporal}, have improved counterfactual quality by integrating shapelets and temporal rules to guide generation. However, these methods face challenges, including the significant processing time required for shapelet extraction and temporal rule mining. TimeX, proposed by \cite{filali2022mining}, extends wCF by introducing a dynamic barycenter average loss term, aiming to produce more contiguous counterfactual explanations.

Among existing methods, Native Guide (NG) \cite{delaney2021instance} leverages the explanation weight vector from class activation mapping (CAM) and the nearest unlike neighbor (NUN) to generate counterfactual solutions. However, NG’s ability to produce high-sparsity counterfactuals depends on the availability of a reliable explanation weight vector. Similarly, \cite{li2022motif} proposed using pre-mined shapelets as the most distinguishable sequences, replacing parts of the original instance with those from the NUN to create new counterfactuals. Nevertheless, the time required for shapelet extraction and the reliance on a single shapelet can limit the method’s ability to consistently generate valid counterfactuals.

More recently, the CELS model \cite{li2023cels} was introduced to address these limitations by incorporating optimization techniques guided by learned saliency maps. This approach identifies the most important time steps to perturb, ensuring more effective generation of counterfactual explanations. However, while CELS showed promise in sparsity and proximity, the validity of the counterfactual explanations is not 100\% guaranteed. 

In this paper, we present an enhanced approach building upon CELS, addressing this limitation by removing mask normalization for more informative and valid counterfactual explanations. Through extensive experimentation on benchmark datasets, we demonstrate that our approach outperforms CELS, achieving higher validity and producing more informative explanations.

The rest of this paper is organized as follows: Section 2 introduces the preliminaries. Section 3 describes the details of the background work CELS. Section 4 describes our proposed method Info-CELS in detail. We present the experimental study in Section 5. Finally, we conclude our work in Section 6.

% Our paper contributions are summarized below:

% \begin{compactenum}
%     \item We present an enhanced method Info-CELS for generating .
%     \item We exploit our learned saliency maps not only to provide intuitive explanations for the reason behind the decision made by the time series classifier on the instance of interest $\mathbf{x}$ but also to explore the post hoc counterfactual explanation for $\mathbf{x}$.
%     \item We provide a comparative empirical study with state-of-the-art methods on five popular real-life datasets from various domains (Spectro, ECG, motion, and simulated) and show the superiority of our methods over state-of-the-art models. 
% \end{compactenum}

% To the best of our knowledge, this is the first effort to learn a saliency map specifically for the purpose of producing high-quality counterfactual explanations of the time series classification problems. 
\section{Preliminaries}\label{sec: pro}
\subsection{Problem Statement}
Given an input time series $\mathbf{x}$ and a classifier $f$ with a prediction $z = f(\mathbf{x})$, a counterfactual explanation seeks to generate a counterfactual instance $\mathbf{x}'$ that minimally modifies $\mathbf{x}$ so that the predicted class changes to a desired label $z'$, where $z' \neq z$. 
The goal is to find a perturbation $\delta$ such that:
\begin{equation}
    f(\mathbf{x}) = z, \quad \mathbf{x}' = \mathbf{x} + \delta \quad \text{such that} \quad f(\mathbf{x}') = z'.
\end{equation}
Here, $\delta$ represents the minimal modification applied to $\mathbf{x}$ to change the model's prediction from $z$ to $z'$.

We refer to $\mathbf{x}$ as the original query instance and $\mathbf{x}'$ as the counterfactual instance (or counterfactual). In binary classification, $z'$ is the class different from $z$, while in multi-class classification, $z'$ is typically chosen as the class with the second-highest probability.
\subsection{Optimal properties for a good counterfactual explanation}\label{sec: pro}
Given an input instance $\mathbf{x}$ and the model's prediction $z$, a counterfactual instance $\mathbf{x}'$ represents the minimal change needed to alter the predicted class to a desired label $z'$. This counterfactual instance is defined as the counterfactual explanation for the input class of interest sample. According to the recent literature on counterfactual explanations for various data modalities \cite{delaney2021instance}, \cite{mothilal2020explaining}, \cite{filali2022mining}, we gathered a comprehensive extended list of key properties of ideal counterfactual explanations for time series data, namely:  

\begin{itemize}
    \item \textbf{Validity}: The prediction result of the to-be-explained model $f$ on the counterfactual instance $\mathbf{x}'$ needs to be different from the prediction result of the to-be-explained model $f$ on the query instance $\mathbf{x}$ (i.e., if $f(\mathbf{x})=z$ and $f(\mathbf{x}')=z'$, then $z \neq z'$).

    \item \textbf{Proximity}: Intuitively, the to-be-explained original query instance of interest needs to be close to the generated counterfactual instance, which means the distance between $\mathbf{x}$ and $\mathbf{x}'$ should be minimal. Proximity can be specified by a distance metric such as the L1 distance. 
    
    \item \textbf{Sparsity}: How many features does a user need to change to transition to the counterfactual class? Intuitively, a counterfactual example will be more feasible if it makes changes to fewer number features. In the case of time series data, the perturbation $\delta$ changing the original query instance $\mathbf{x}$ into $\mathbf{x}'$ should be sparse, which means fewer data points that have been changed to obtain the counterfactual explanation is preferred.
    
    \item \textbf{Contiguity}: The counterfactual instance $\mathbf{x}'$ needs to be perturbed with fewer independent contiguous segments, ensuring a more semantically meaningful solution.
    
    \item \textbf{Interpretable\footnote{Also referred to as plausibility (e.g., \cite{delaney2021instance}.)}}: The counterfactual $\mathbf{x}'$ needs to be in-distribution. We consider a counterfactual instance $\mathbf{x}'$ interpretable if it lies close to the model’s training data distribution. $\mathbf{x}'$ should be an inlier with respect to the training dataset (query instances) and an inlier to the counterfactual class. 
    
    % \item \textbf{Model-agnosticism}: Ideally, the counterfactual explanation model should produce a solution independent of the classification model $f$. 
\end{itemize}

\section{Background Works: CELS}
Recently, the CELS model \cite{li2023cels}, for the first time, introduced the idea of generating counterfactual explanations for time series classification through learned saliency maps. Here, we describe the methodology, that this paper uses and expands to enable us to provide 100\% validity and informative counterfactual explanations. 

Given a set of $N$ time series $\mathcal{D}=$ $\left\{\mathbf{x}_1, \ldots, \mathbf{x}_N\right\}$ and a time series classification model $f: \mathbb{X} \rightarrow \mathcal{Y}$, where $\mathbb{X} \in \mathbb{R}^T$ is the $T$-dimensional feature space and $\mathcal{Y}=\{1,2, \ldots, C\}$ is the label space with $C$ classes. Let us consider an instance-of-interest $\mathbf{x} \in \mathcal{D}$, a time series with $T$ time steps along with a predicted probability distribution $\hat{\mathbf{y}}=f(\mathbf{x})$ over $C$ classes where $\hat{\mathbf{y}} \in[0,1]^C$ and $\sum_{i=1}^C \hat{\mathbf{y}}_i=1$. The top predicted class $z$ where $z=\operatorname{argmax} \hat{\mathbf{y}}$ has predicted confidence $\hat{\mathbf{y}}_{\mathbf{z}}$ is the class of interest for which we seek a counterfactual explanation.

The goal is to learn a saliency map $\boldsymbol{\theta} \in$ $[0,1]^T$ for the class of interest $z$ where each element represents the importance of a corresponding time step at the very first step. Values in $\boldsymbol{\theta}$ close to 1 indicate strong evidence for class $z$, while values in $\boldsymbol{\theta}$ close to 0 indicate no importance. We assume the importance of a time step $t$ should reflect the expected scale of the change of $P\left(\hat{\mathbf{y}}_z \mid \mathbf{x}\right)$ when $\mathbf{x}$ is perturbed: $\left|P\left(\hat{\mathbf{y}}_z \mid \mathbf{x}\right)-P\left(\hat{\mathbf{y}}_z \mid \mathbf{x}' \right)\right|$, where $\mathbf{x}'$ is perturbed version of $\mathbf{x}$. Then we design a perturbation function based on the learned saliency map to generate a counterfactual explanation $\mathbf{x}'$ for the class of interest $z$. A successfully learned saliency map $\boldsymbol{\theta}$ will assign high values to the regions in a time series where the perturbation function would shift the model's predictions away from $z$. We define the instance of interest $\mathbf{x}$ for which we seek a counterfactual explanation as the original query instance and the perturbed result $\mathbf{x}'$ as the counterfactual instance or counterfactual.

\begin{figure*}[ht!]
\centering
\includegraphics[scale = .35]{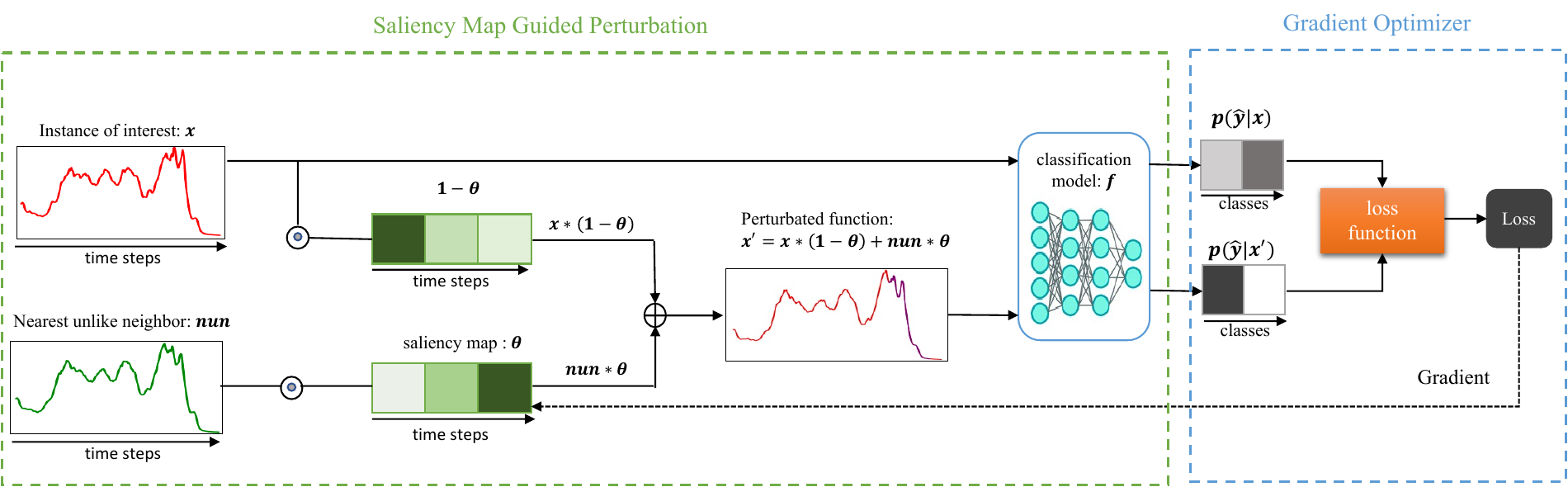}
\caption{Counterfactual explanation for time series data via learned saliency maps (CELS)}
\label{Fig:mainfig}
\end{figure*}

CELS \cite{li2023cels} contains three key components that work together: (1) the Nearest Unlike Neighbor Replacement strategy learns which time series from the background dataset $\mathcal{D}$ are the best for replacement-based perturbation. (2) the Learning Explanation aims to learn a saliency map $\boldsymbol{\theta}$ for the instance of interest by highlighting the most important time steps that provide evidence for a model prediction. (3) the Saliency Map Guided Counterfactual Perturbation function which perturbs the instance of interest $\mathbf{x}$ to force the perturbation shift the model's prediction away from the class of interest $z$. Figure \ref{Fig:mainfig} shows the pipeline of generating a counterfactual explanation for a given instance of interest $\mathbf{x}$ using CELS.

\subsection{Replacement Strategy: Nearest Unlike Neighbor (NUN)}
Perturbing the values of the instance-of-interest $\mathbf{x}$ introduces changes and produces a synthetic time series $\mathbf{x}'$, intended to serve as a counterfactual explanation for $\mathbf{x}$. However, determining which time steps to modify and the extent of those modifications can have significant implications. Drastic alterations to $\mathbf{x}$ often lead to perturbations that are outside the distribution of the data, rendering the predictions unreliable. To address this issue, NG \cite{delaney2021instance} proposes a solution by generating in-distribution perturbations through the replacement of time steps with those from the nearest unlike neighbor instances in a background dataset $\mathcal{D}$. Drawing inspiration from NG, CELS adopts a similar approach by selecting the nearest unlike neighbor (nun) time series from the training dataset to guide the perturbation of the instance of interest $\mathbf{x}$. The nearest unlike neighbor is defined as finding an instance $nun$ from a different class $z'$, with the minimum distance to $\mathbf{x}$, using a specific distance metric. In binary classification cases, $z'$ is chosen as the class different from $z$. In a multi-class classification setting, $z'$ is designated as the class with the second-highest probability.  This strategy offers insights into the rationale behind the model predictions and facilitates the generation of counterfactual explanations, showcasing the minimal changes necessary to modify the prediction outcome \cite{delaney2021instance}. 

\subsection{Counterfactual Perturbation}
A saliency map is derived for the class of interest $z$, the top predicted class. Then derived saliency map $\boldsymbol{\theta}$ can be used to generate the counterfactual explanation. CELS perturbs the instance of interest $\mathbf{x}$ by performing a time-step replacement using the nearest unlike neighbor $nun$ and the saliency map $\boldsymbol{\theta}$ to guide the counterfactual explanation. Equation \ref{eq:perturb}
shows the counterfactual perturbation function. 

Specifically, CELS learns a perturbation function $\Phi: \mathbb{R}^{1 \times T} \rightarrow \mathbb{R}^{1 \times T}$ where the key component of $\Phi$ is a one-dimensional parameterized matrix $\boldsymbol{\theta} \in$ $[0,1]^{1 \times T}$. An element $\boldsymbol{\theta}_t$ of this matrix represents the importance value of time steps $t$ for class $z$. Values of $\boldsymbol{\theta}$ close to 1 indicate strong evidence for the class $z$. Values close to 0 imply no importance. Therefore, ideally, if $\boldsymbol{\theta}_t = 0$, which means that the time step $t$ is not important for the prediction decision on class $z$, then the original time step $\mathbf{x}_t$ remains unchanged. In other words, if $\boldsymbol{\theta}_t = 1$, which means that the time steps $t$ provide important evidence for the prediction decision on class $z$, then $\mathbf{x}_t$ is replaced with the corresponding time step of nearest unlike neighbor $nun$. This way, CELS learns the degree of sensitivity of each time step for $\mathbf{x}$. The perturbation function $\Phi\left(\mathbf{x}; \boldsymbol{\theta}\right)$ generates the counterfactual explanation $\mathbf{x}'$ by performing time step interpolation between the original time steps of $\mathbf{x}$ and the replacement series $nun$.

\begin{equation}
\label{eq:perturb}
    \mathbf{x}'=  \mathbf{x} \odot \left(1-\boldsymbol{\theta}\right) +  nun \odot \boldsymbol{\theta}
\end{equation}

\subsection{Learning Explanation}
Saliency values $\mathbb{\theta} \in \mathbb{R}^{1 \times T}$ are learned using a novel loss function (Equation \ref{eq: total loss}) including three key components. 

First, a $L_{\operatorname{Max}}$ is designed to promote the perturbation function to generate high-validity counterfactual instances. Specifically, $L_{\operatorname{Max}}$ is incorporated to maximize the class probability of class $z'$ predicted by the classification model $f$ for the perturbated counterfactual instance:

\begin{equation}\label{eq: max}
    L_{\operatorname{Max}}= \left(1-P\left(\hat{\mathbf{y}}_{z'} \mid \mathbf{x}'\right)\right)
\end{equation}

Next, to encourage simple explanations with minimal salient time steps, the $L_{\text {Budget }}$ loss is introduced. This component promotes the values of $\boldsymbol{\theta}$ to be as small as possible, where intuitively, values that are not important should be close to 0 according to the defined problem:

\begin{equation}\label{eq: Budget}
    L_{\text {Budget }}=\frac{1}{T} \sum_{t=1}^T\boldsymbol{\theta}_{ t}
\end{equation}

The saliency map is then encouraged to be temporally coherent, where neighboring time steps should generally have similar importance. To achieve this coherence, a time series regularizer $L_{\text {TReg }}$ is introduced, minimizing the squared difference between neighboring saliency values:

\begin{equation}\label{eq: treg}
    L_{\mathrm{TReg}}=\frac{1}{T} \sum_{t=1}^{T-1}\left(\boldsymbol{\theta}_{t}-\boldsymbol{\theta}_{t+1}\right)^2
\end{equation}

Finally, all the loss terms are summed, with $L_{\operatorname{Max}}$ being scaled by a $\lambda$ coefficient to balance the three components in counterfactual explanation behavior. Minimizing the total loss in Equation \ref{eq: total loss} leads to simple saliency maps and minimum change in the counterfactual explanation.
\begin{equation}
\label{eq: total loss}
    L(P(\hat{\mathbf{y}} \mid \mathbf{x}) ; \boldsymbol{\theta})=\lambda *L_{\mathrm{Max}}+L_{\mathrm{Budget}}+L_{\mathrm{TReg}}
\end{equation}

\begin{algorithm}
\caption{CELS Algorithm\label{alg:alg1}} 
\textbf{Inputs:} Instance of interest $\mathbf{x}$, pretrained time series classification model $f$, background dataset $\mathcal{D}$, the total number of time steps $T$ of the instance, a threshold $k$ to normalize $\boldsymbol{\theta}$ at the end of learning.\\
\textbf{Output:} Saliency map $\boldsymbol{\theta}$ for instance of interest $\mathbf{x}$, counterfactual explanation $\mathbf{x}'.$
\begin{algorithmic}[1]
  \State $\boldsymbol{\theta}$ $\leftarrow$ random\_uniform(low = 0,high = 1)
  \State $nun, z'$ $\leftarrow$ Nearest Unlike Neighbor($\mathbf{x}$, $\mathcal{D}$, $f$) 
  \For{epoch $\leftarrow$ 1 to epochs}
      \State $\mathbf{x}'= \mathbf{x} \odot \left(1-\boldsymbol{\theta}\right) +  nun \odot \boldsymbol{\theta}$
      \State $ L_{\operatorname{Max}}= \left(1-P\left(\hat{\mathbf{y}}_{z'} \mid \mathbf{x}'\right)\right)$
      \State $L_{\mathrm{Budget}}=\frac{1}{T} \sum_{t=1}^T\boldsymbol{\theta}_{ t}$
      \State $ L_{\mathrm{TReg}}=\frac{1}{T} \sum_{t=1}^{T-1}\left(\boldsymbol{\theta}_{t}-\boldsymbol{\theta}_{t+1}\right)^2$
      \State $L(P(\hat{\mathbf{y}} \mid \mathbf{x}) ; \boldsymbol{\theta})=\lambda *L_{\mathrm{Max}}+L_{\mathrm{Budget}}+L_{\mathrm{TReg}}$
      \State $\boldsymbol{\theta} \leftarrow \boldsymbol{\theta} - \eta \nabla J(\boldsymbol{\theta})$
      \State clamp($\boldsymbol{\theta}$, low = 0, high = 1)   
      \Comment{clamp ensures that $\theta$ is within the (0,1) range }
  \EndFor
  \State \textcolor{blue}{$\boldsymbol{\theta}[i] \leftarrow (\boldsymbol{\theta}[i]  > k) \, ? \, 1 \, : \, 0$}
  \Comment{\textcolor{blue}{normalize the saliency map to get the final counterfactual explanation}
  \State \textcolor{blue}{$\mathbf{x}'= \mathbf{x} \odot \left(1-\boldsymbol{\theta}\right) +  nun \odot \boldsymbol{\theta}$}}
  \Comment{\textcolor{blue}{recompute the counterfactual explanation $\mathbf{x}'$ using the normalized saliency map}}
  \State \Return {$\boldsymbol{\theta}$, $\mathbf{x}'$}
  
\end{algorithmic}
\end{algorithm}
\vspace{-20pt}

\section{Informative CELS (Info-CELS)}
In the CELS algorithm, a saliency map normalization step is performed before generating the final counterfactual explanation $\mathbf{x}'$ (see Algorithm \ref{alg:alg1}, line 11). Specifically, after several optimization epochs, the learned saliency map $\boldsymbol{\theta}$ is normalized using a threshold $k$ (see Algorithm \ref{alg:alg1}, line 11). For each $\boldsymbol{\theta}[i]$, if $\boldsymbol{\theta}[i] > k$, it is set to 1; otherwise, it is set to 0, indicating whether the corresponding time step is considered important for perturbation. Subsequently, the final counterfactual explanation $\mathbf{x}'$ is recalculated based on the normalized saliency map (see Algorithm \ref{alg:alg1}, line 12). Initially, this normalization aimed to minimize perturbations to the original query instance, thus producing counterfactual explanations with higher sparsity. However, this normalization step imposes limitations on the validity of the counterfactual explanation. 

\begin{figure*}[ht!]
        \centering
        \begin{subfigure}[b]{0.40\textwidth}
            \centering
            \includegraphics[width=\textwidth]{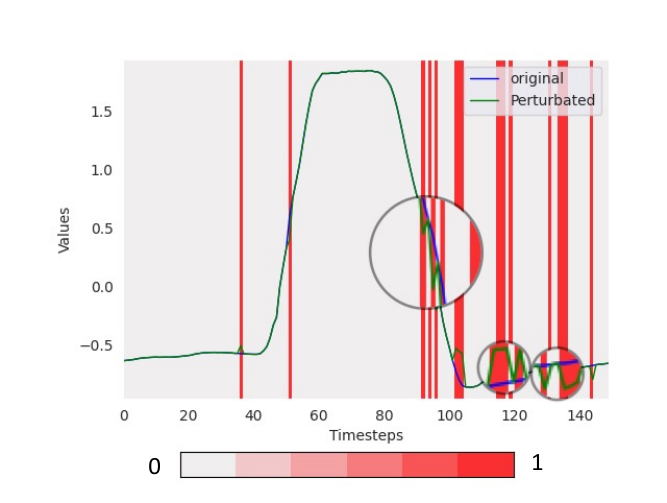}
            \caption[]%
            {{\small CELS GunPoint}}    
            \label{fig:celsgun}
        \end{subfigure}
        % \hfill
        \begin{subfigure}[b]{0.40\textwidth}  
            \centering 
            \includegraphics[width=\textwidth]{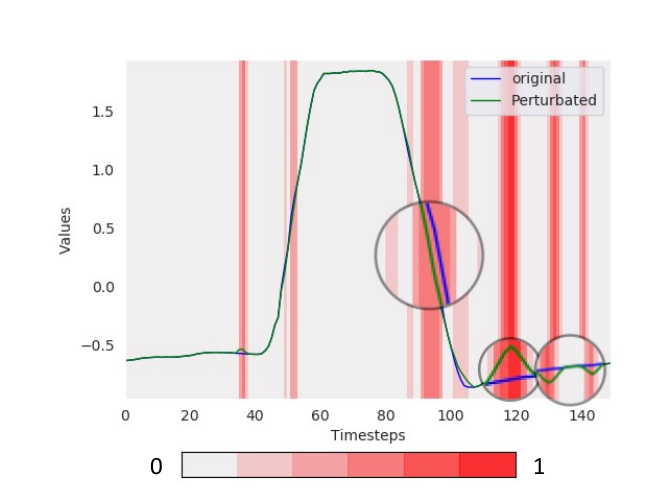}
            \caption[]%
            {{\small Info-CELS GunPoint}}    
            \label{fig:infocelsgun}
        \end{subfigure}
      
        % \vskip\baselineskip
        \begin{subfigure}[b]{0.40\textwidth}   
            \centering 
            \includegraphics[width=\textwidth]{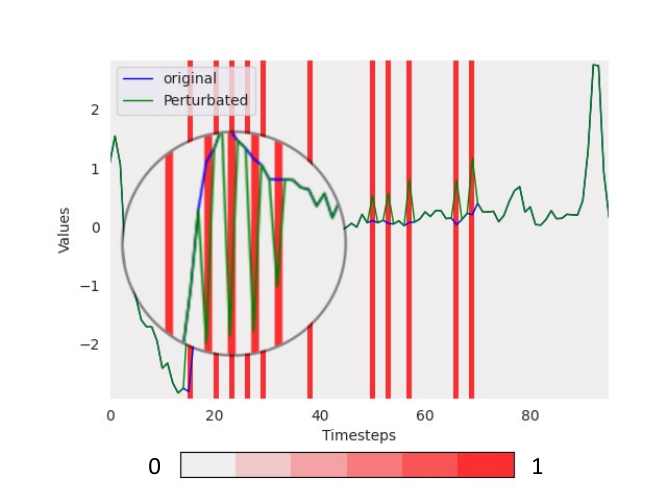}
            \caption[]%
            {{\small CELS ECG200}}    
            \label{fig:celsecg}
        \end{subfigure}
        % \hfill
        \begin{subfigure}[b]{0.40\textwidth}   
            \centering 
            \includegraphics[width=\textwidth]{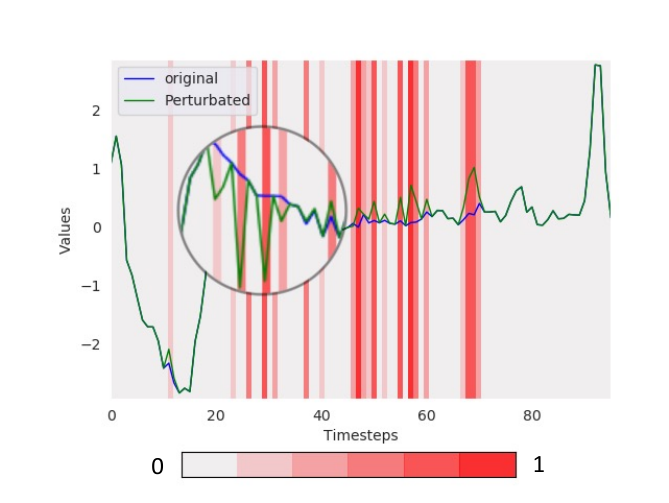}
            \caption[]%
            {{\small Info-CELS ECG200}}    
            \label{fig:infocelsecg}
        \end{subfigure}
        \caption[]%
        {\small Counterfactual explanations obtained from CElS and Info-CELS for the ECG200 and GunPoint datasets. (The x-axis in each figure represents the time steps, indicating the progression of time in the time series data. The y-axis denotes the values corresponding to each time step, showing the changes in the data over time. Below the figure, the 1-dimensional saliency map highlights the importance scores for each time step, ranging between 0 and 1. A score of 1 means the time step is most influential in the model's decision-making process.)}
        \label{fig:cfexample}
\end{figure*}

To illustrate, Figures \ref{fig:celsgun} and \ref{fig:celsecg} display the normalized saliency maps and the counterfactual explanations generated by CELS for instances from the GunPoint and ECG datasets, respectively. It becomes apparent that the normalization process in CELS, while intended to reduce perturbations and increase sparsity, inadvertently introduces noise into the counterfactual instances, characterized by more significant abrupt changes. This noise can obscure the underlying rationale of the model's decision-making process, potentially undermining the reliability and validity of the explanations.

To address this issue, we introduce Info-CELS, which omits the normalization step outlined in lines 12-13 of Algorithm \ref{alg:alg1}. By removing this step, Info-CELS guides perturbations based on the learned saliency map without normalization, aiming to maintain smoother representations of the perturbed instances. The comparison in Figure \ref{fig:cfexample} illustrates the difference between counterfactual explanations generated by CELS and Info-CELS. Notably, Info-CELS produces counterfactuals with smoother transitions and fewer abrupt changes, indicating reduced noise and enhanced interpretability.

Furthermore, our experiments reveal that Info-CELS achieves 100\% validity while maintaining comparable levels of sparsity and proximity to CELS-generated explanations (see section \ref{sec: eva res}). In summary, Info-CELS enhances the validity and interpretability of counterfactual explanations by mitigating the risk of introducing unnecessary noise, thereby improving trustworthiness. Further experimental study and findings will be elaborated in the subsequent section.
% \vspace{-20pt}

\section{Experimental Study}
In this section, we outline our experimental design and discuss our findings. We conducted our experiments on publicly available univariate time series data sets from the University of California at Riverside (UCR) Time Series Classification Archive \cite{dau2019ucr}. In particular, we selected seven popular real-life datasets from various domains (Spectro, ECG, Motion, Simulated, Image, and Sensor) from the UCR archive that demonstrate good classification accuracy on state-of-the-art classifiers as reported in \cite{ismail2019deep}. This ensures the quality of our generated counterfactual instances. The time series classes are evenly sampled across the training and testing partitions. Table~\ref{tab:UCR} shows all the details of the data set. The original test set size of TwoLeadECG and CBF are 1139 and 900, considering the computation time, we randomly choose 100 samples from each test set to learn the counterfactual explanation. 
% \vspace{-25pt}
\begin{table*}[!h]
\centering
\caption{UCR Datasets Metadata}
\begin{tabular}{|c||c|c|c|c|c|c|c|}
\hline
\label{tab:UCR}
ID & Dataset Name & $\mathcal{C}$  & $\mathcal{L}$ & DS train size & DS test size & Type\\ \hline
0  & Coffee       & 2  & 286       & 28    &28   &SPECTRO\\ \hline
1  & GunPoint     & 2  & 150       & 50    &150  &MOTION\\ \hline
2  & ECG200       & 2  & 96        & 100   &100  &ECG\\ \hline
3  & TwoLeadECG   & 2  & 82        & 23    &100  &ECG\\ \hline
4  & CBF          & 3  & 128       & 30    &100  &SIMULATED\\ \hline
5  & BirdChicken  & 2  & 512    & 20    &20 &IMAGE\\ \hline
6  & Plane       & 7  & 144   & 105   &105 &SENSOR\\ \hline
\end{tabular}

\footnotesize {$\mathcal{C}$: number of classes; $\mathcal{L}$: time series length, DS: dataset}
\end{table*}
% \vspace{-25pt}

\subsection{Baseline Methods}
We evaluated the Info-CELS model with CELS and the other three baselines, Alibi, Native guide counterfactual (NG), and SG.
\begin{itemize}
    \item \textbf{CELS \cite{li2023cels}}: a novel counterfactual explanation method for time series classification by learning saliency maps to guide the perturbation, which is the first attempt to not only to provide intuitive explanations for the reason behind the decision made by the time series classifier on the instance of interest $\mathbf{x}$ but also to explore the post hoc counterfactual explanation for $\mathbf{x}$.
    
    \item \textbf{Alibi Counterfactual (ALIBI):} ALIBI follows the work
of Wachter et al. \cite{wachter2017counterfactual}, which constructs counterfactual explanations by optimizing an objective function, 
     \begin{equation}
        L = L_{pred} + L_{L1},
    \end{equation}
     where $L_{pred}$ guides the search towards points $\textbf{x}^{\prime}$ which would change the model prediction and $L_{L1}$ ensures that $\textbf{x}^{\prime}$ is close to $\textbf{x}$. 
    
    \item \textbf{Native guide counterfactual (NG):} NG \cite{delaney2021instance} is an instance-based counterfactual method that uses the explanation weight vector and the in-sample counterfactual to generate a new counterfactual solution. 

    \item \textbf{Shapelet-guied counterfactual (SG):} SG \cite{li2022sg} is the extension based on wCF which introduces a new shapelet loss term (see Equation~\ref{eq:lossproto}) to enforce the counterfactual instance $\mathbf{x}'$ to be close to the prominent shapelet $\mathbf{x_{shapelet}}$ extracted from the training set.
    \begin{equation}
    L_{shapelet} = d(\mathbf{x}',\mathbf{x_{shapelet}})
    % L_{shapelet} = d(x_{cf} - x_{shapelet})
    \label{eq:lossproto} 
    \end{equation}
    \begin{equation}
    L = L_{pred} + L_{L1} + L_{shapelet} 
    \label{eq:newloss} 
    \end{equation}
\end{itemize}

\subsection{Implementation Details}
For each dataset, we evaluated all the aforementioned counterfactual baselines on the same fully convolutional neural network (FCN), which was originally proposed by Wang et al. \cite{wang2017time} and implemented by Fawaz et al. \cite{ismail2019deep}. Each dataset from the UCR archive comes with a pre-defined split ratio. We train each model only on the training data and generate counterfactuals for all test instances. We optimize Info-CELS using ADAM \cite{kingma2014adam} with a learning rate of 0.1 and train for 1000 epochs, we also set an early stopping to end the optimization process earlier when a satisfactory level of convergence on the loss function is achieved. The $\lambda$ we used in our experiments is set to 1.

\subsection{Evaluation metrics \label{Sec:ExpResults}}
To evaluate our proposed method, we first compare Info-CELS with the other four baselines in terms of validity by evaluating the label flip rate for the prediction of the counterfactual explanation result and the target probability, we computed the flip rate following the formula of Equation~\ref{equ:flipr}.
\begin{equation}\label{equ:flipr}
    flip\_rate = \frac{num\_flipped}{num\_testsample},
\end{equation}

where we denote the num\_flipped as the number of generated counterfactual explanation samples that result in a different prediction label compared to the original input, and the num\_test\_samples is the total number of samples in the test dataset used to generate these counterfactual explanations. According to the validity property as we mentioned in section \ref{sec: pro}, a counterfactual instance should result in a prediction that is different from the prediction of the original instance. Therefore, the flip rate directly aligns with this validity property by quantifying how often the generated counterfactual instances meet this criterion. A high flip rate 1 indicates that all the generated counterfactual instances are valid.

The target class probability measures the confidence level of the prediction for the counterfactual explanation. Specifically, it assesses how likely the model is to predict the counterfactual instance as belonging to the target class. The closer this probability is to 1, the more confident and desirable the prediction. This metric provides an additional layer of evaluation by ensuring that not only are the counterfactual instances valid (as measured by the flip rate) but they are also predicted with high confidence. The results of these evaluations are presented in Table \ref{tab: flip rate and target probability}.

Next, we compare the L1 distance and sparsity level to evaluate the proximity and sparsity. L1 distance is defined in Equation~\ref{equ:L1}. It measures the closeness between the generated counterfactual $\mathbf{x}'$ and the original instance of interest $\mathbf{x}$, a smaller L1 distance is preferred. 

\begin{equation}
   L1 \ distance = \sum_{t=1}^T(|\mathbf{x_{t}}- \mathbf{x}_{t}'|), 
    \label{equ:L1} 
\end{equation}

For the sparsity level, the highest sparsity is an indicator that the number of time series data points perturbations made in $\mathbf{x}$ to achieve $\mathbf{x}'$ is minimal. Therefore, a higher sparsity level is desirable. The equations designed by \cite{filali2022mining} are shown in \ref{equ:spar}-\ref{equ:func}, where T represents the total number of time steps (length) of the time series data.

\begin{equation}\label{equ:spar}
    Sparsity = 1- \frac{\sum_{t=1}^{T}g({\mathbf{x}'_t}, \mathbf{x}_t)}{T} 
\end{equation}
\begin{equation}\label{equ:func}
    g(x, y)=\left\{\begin{array}{ll}
    1, & \text { if } x \neq y \\
    0, & \text { otherwise }
    \end{array}\right. 
\end{equation} 

\subsection{Evaluation results} \label{sec: eva res}

% \vspace{-25pt}
\begin{table*}[htbp]
\centering
\caption{Comparing the performances among different counterfactual explanation models in terms of flip rate and the target probability (The winner is bolded).}
\label{tab: flip rate and target probability}
\begin{tabular}{|l|
>{\columncolor[HTML]{C0C0C0}}l 
>{\columncolor[HTML]{C0C0C0}}l 
>{\columncolor[HTML]{C0C0C0}}l 
>{\columncolor[HTML]{C0C0C0}}l 
>{\columncolor[HTML]{C0C0C0}}l |
>{\columncolor[HTML]{EFEFEF}}l 
>{\columncolor[HTML]{EFEFEF}}l 
>{\columncolor[HTML]{EFEFEF}}l 
>{\columncolor[HTML]{EFEFEF}}l 
>{\columncolor[HTML]{EFEFEF}}l |}
\hline
 & \multicolumn{5}{c|}{\cellcolor[HTML]{C0C0C0}\textbf{Flip rate}} & \multicolumn{5}{c|}{\cellcolor[HTML]{EFEFEF}\textbf{Target probability}} \\ \hline
Dataset & \multicolumn{1}{l|}{\cellcolor[HTML]{C0C0C0}NG} & \multicolumn{1}{l|}{\cellcolor[HTML]{C0C0C0}ALIBI} & \multicolumn{1}{l|}{\cellcolor[HTML]{C0C0C0}SG} & \multicolumn{1}{l|}{\cellcolor[HTML]{C0C0C0}CELS} & Info-CELS & \multicolumn{1}{l|}{\cellcolor[HTML]{EFEFEF}NG} & \multicolumn{1}{l|}{\cellcolor[HTML]{EFEFEF}ALIBI} & \multicolumn{1}{l|}{\cellcolor[HTML]{EFEFEF}SG} & \multicolumn{1}{l|}{\cellcolor[HTML]{EFEFEF}CELS} & Info-CELS \\ \hline
Coffee & \multicolumn{1}{l|}{\cellcolor[HTML]{C0C0C0}\textbf{1.0}} & \multicolumn{1}{l|}{\cellcolor[HTML]{C0C0C0}0.79} & \multicolumn{1}{l|}{\cellcolor[HTML]{C0C0C0}0.61} & \multicolumn{1}{l|}{\cellcolor[HTML]{C0C0C0}0.786} & \textbf{1.0} & \multicolumn{1}{l|}{\cellcolor[HTML]{EFEFEF}0.67} & \multicolumn{1}{l|}{\cellcolor[HTML]{EFEFEF}0.74} & \multicolumn{1}{l|}{\cellcolor[HTML]{EFEFEF}0.60} & \multicolumn{1}{l|}{\cellcolor[HTML]{EFEFEF}0.62} & \textbf{0.9} \\ \hline
GunPoint & \multicolumn{1}{l|}{\cellcolor[HTML]{C0C0C0}\textbf{1.0}} & \multicolumn{1}{l|}{\cellcolor[HTML]{C0C0C0}0.95} & \multicolumn{1}{l|}{\cellcolor[HTML]{C0C0C0}0.92} & \multicolumn{1}{l|}{\cellcolor[HTML]{C0C0C0}0.680} & \textbf{1.0} & \multicolumn{1}{l|}{\cellcolor[HTML]{EFEFEF}0.63} & \multicolumn{1}{l|}{\cellcolor[HTML]{EFEFEF}0.87} & \multicolumn{1}{l|}{\cellcolor[HTML]{EFEFEF}0.86} & \multicolumn{1}{l|}{\cellcolor[HTML]{EFEFEF}0.59} & \textbf{0.97} \\ \hline
ECG200 & \multicolumn{1}{l|}{\cellcolor[HTML]{C0C0C0}\textbf{1.0}} & \multicolumn{1}{l|}{\cellcolor[HTML]{C0C0C0}0.88} & \multicolumn{1}{l|}{\cellcolor[HTML]{C0C0C0}0.93} & \multicolumn{1}{l|}{\cellcolor[HTML]{C0C0C0}0.820} & \textbf{1.0} & \multicolumn{1}{l|}{\cellcolor[HTML]{EFEFEF}0.79} & \multicolumn{1}{l|}{\cellcolor[HTML]{EFEFEF}0.82} & \multicolumn{1}{l|}{\cellcolor[HTML]{EFEFEF}\textbf{0.93}} & \multicolumn{1}{l|}{\cellcolor[HTML]{EFEFEF}0.68} & 0.86 \\ \hline
TwoLeadECG & \multicolumn{1}{l|}{\cellcolor[HTML]{C0C0C0}\textbf{1.0}} & \multicolumn{1}{l|}{\cellcolor[HTML]{C0C0C0}0.87} & \multicolumn{1}{l|}{\cellcolor[HTML]{C0C0C0}\textbf{1.0}} & \multicolumn{1}{l|}{\cellcolor[HTML]{C0C0C0}0.6} & \textbf{1.0} & \multicolumn{1}{l|}{\cellcolor[HTML]{EFEFEF}0.80} & \multicolumn{1}{l|}{\cellcolor[HTML]{EFEFEF}0.83} & \multicolumn{1}{l|}{\cellcolor[HTML]{EFEFEF}\textbf{0.99}} & \multicolumn{1}{l|}{\cellcolor[HTML]{EFEFEF}0.57} & 0.84 \\ \hline
CBF & \multicolumn{1}{l|}{\cellcolor[HTML]{C0C0C0}0.56} & \multicolumn{1}{l|}{\cellcolor[HTML]{C0C0C0}0.89} & \multicolumn{1}{l|}{\cellcolor[HTML]{C0C0C0}0.81} & \multicolumn{1}{l|}{\cellcolor[HTML]{C0C0C0}0.71} & \textbf{1.0} & \multicolumn{1}{l|}{\cellcolor[HTML]{EFEFEF}0.41} & \multicolumn{1}{l|}{\cellcolor[HTML]{EFEFEF}0.83} & \multicolumn{1}{l|}{\cellcolor[HTML]{EFEFEF}0.78} & \multicolumn{1}{l|}{\cellcolor[HTML]{EFEFEF}0.56} & \textbf{0.86} \\ \hline
{BirdChicken} & \multicolumn{1}{l|}{\cellcolor[HTML]{C0C0C0}{\textbf{1.0}}} & \multicolumn{1}{l|}{\cellcolor[HTML]{C0C0C0}{\textbf{1.0}}} & \multicolumn{1}{l|}{\cellcolor[HTML]{C0C0C0}{0.85}} & \multicolumn{1}{l|}{\cellcolor[HTML]{C0C0C0}{0.5}} & \cellcolor[HTML]{C0C0C0}{\textbf{\textbf{1.0}}} & \multicolumn{1}{l|}{\cellcolor[HTML]{EFEFEF}{0.63}} & \multicolumn{1}{l|}{\cellcolor[HTML]{EFEFEF}{0.81}} & \multicolumn{1}{l|}{\cellcolor[HTML]{EFEFEF}{0.80}} & \multicolumn{1}{l|}{\cellcolor[HTML]{EFEFEF}{0.53}} & \cellcolor[HTML]{EFEFEF}{\textbf{0.96}} \\ \hline
{Plane} & \multicolumn{1}{l|}{\cellcolor[HTML]{C0C0C0}{0.14}} & \multicolumn{1}{l|}{\cellcolor[HTML]{C0C0C0}{0.69}} & \multicolumn{1}{l|}{\cellcolor[HTML]{C0C0C0}{0.87}} & \multicolumn{1}{l|}{\cellcolor[HTML]{C0C0C0}{0.57}} & \cellcolor[HTML]{C0C0C0}{\textbf{1.0}} & \multicolumn{1}{l|}{\cellcolor[HTML]{EFEFEF}{0.16}} & \multicolumn{1}{l|}{\cellcolor[HTML]{EFEFEF}{0.67}} & \multicolumn{1}{l|}{\cellcolor[HTML]{EFEFEF}{\textbf{0.85}}} & \multicolumn{1}{l|}{\cellcolor[HTML]{EFEFEF}{0.4}} & \cellcolor[HTML]{EFEFEF}{0.81} \\ \hline
\end{tabular}
\end{table*}

% \vspace{-25pt}

\subsubsection{Validity evaluation: }
Table \ref{tab: flip rate and target probability} shows the performance comparison among different counterfactual explanation models in terms of flip label rate and the target probability. Based on the results in Table \ref{tab: flip rate and target probability}, Info-CELS consistently outperforms NG, ALIBI, SG, and CELS, achieving a stable 100\% flip label rate with competitive target probability across various datasets. The lower flip rate of CELS is due to noise from mask normalization, as discussed in Section 4. SG's instability in validity may stem from the varying quality of selected shapelets. ALIBI's global loss minimization can make finding optimal perturbations challenging. NG performs well in binary classification but struggles with multi-class cases, likely due to the quality of class activation maps.

The high flip label rate achieved by Info-CELS indicates its effectiveness in generating counterfactual explanations that lead to changes in the predicted labels, thereby providing valuable insights into model behavior and decision-making processes. The stability of achieving a 100\% validity rate across all datasets further underscores the reliability and consistency of our method in producing meaningful explanations. 

Moreover, the consistently high target probability obtained by Info-CELS across all datasets highlights its ability to generate counterfactual explanations that are not only valid but also align closely with the desired outcomes. The superior performance of Info-CELS compared to the other baselines, particularly in terms of target probability, underscores its efficacy in providing accurate and informative counterfactual explanations. This implies that Info-CELS can offer valuable insights with a high confidence level into the decision-making process of complex machine learning models across diverse datasets, ultimately facilitating better understanding and trust in AI systems. 
% \vspace{-15pt}
\begin{figure*}[h!]
    \centering
    \setkeys{Gin}{width=0.5\linewidth}
\subfloat[Sparsity level\label{fig: Sparsity}]{\includegraphics{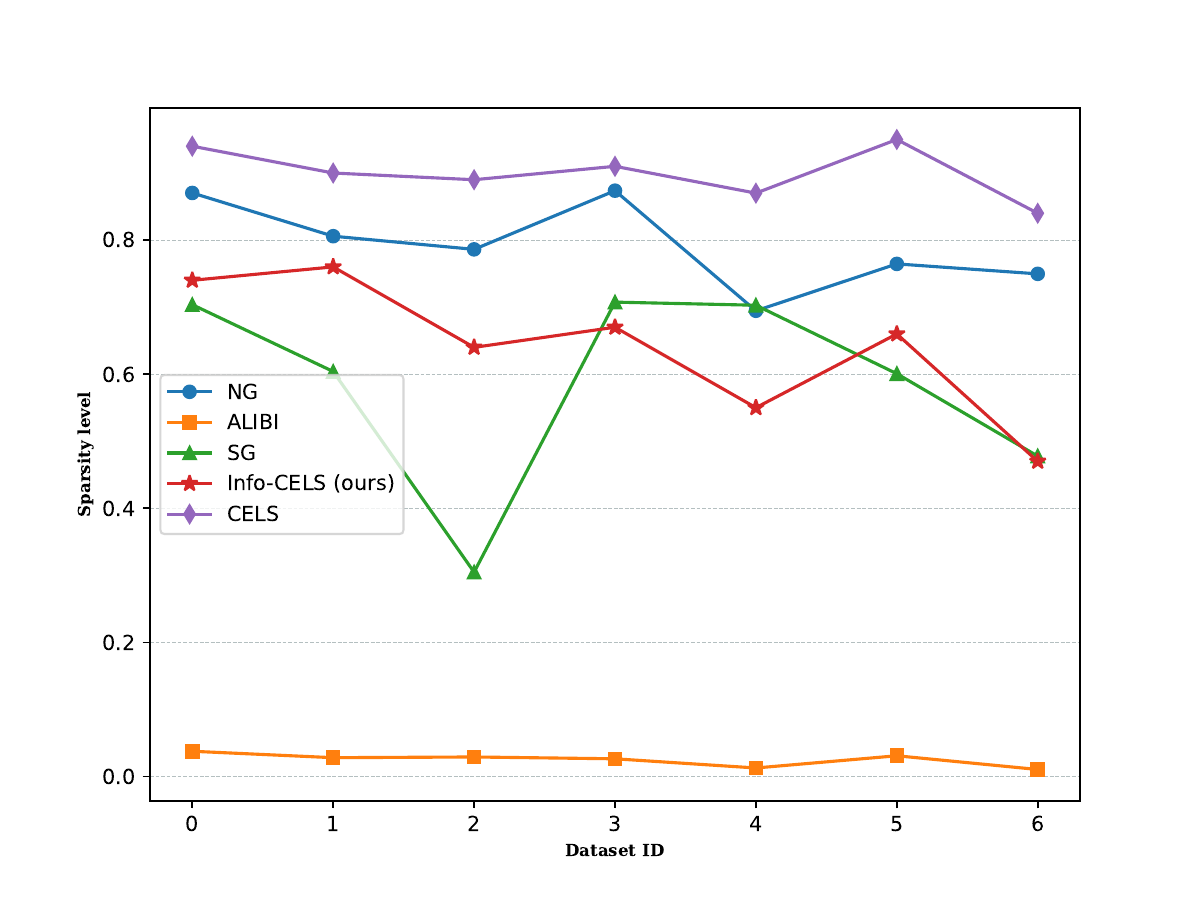}}% 1
\hfill
\subfloat[L1 distance (Proximity) \label{fig: L1}]{\includegraphics{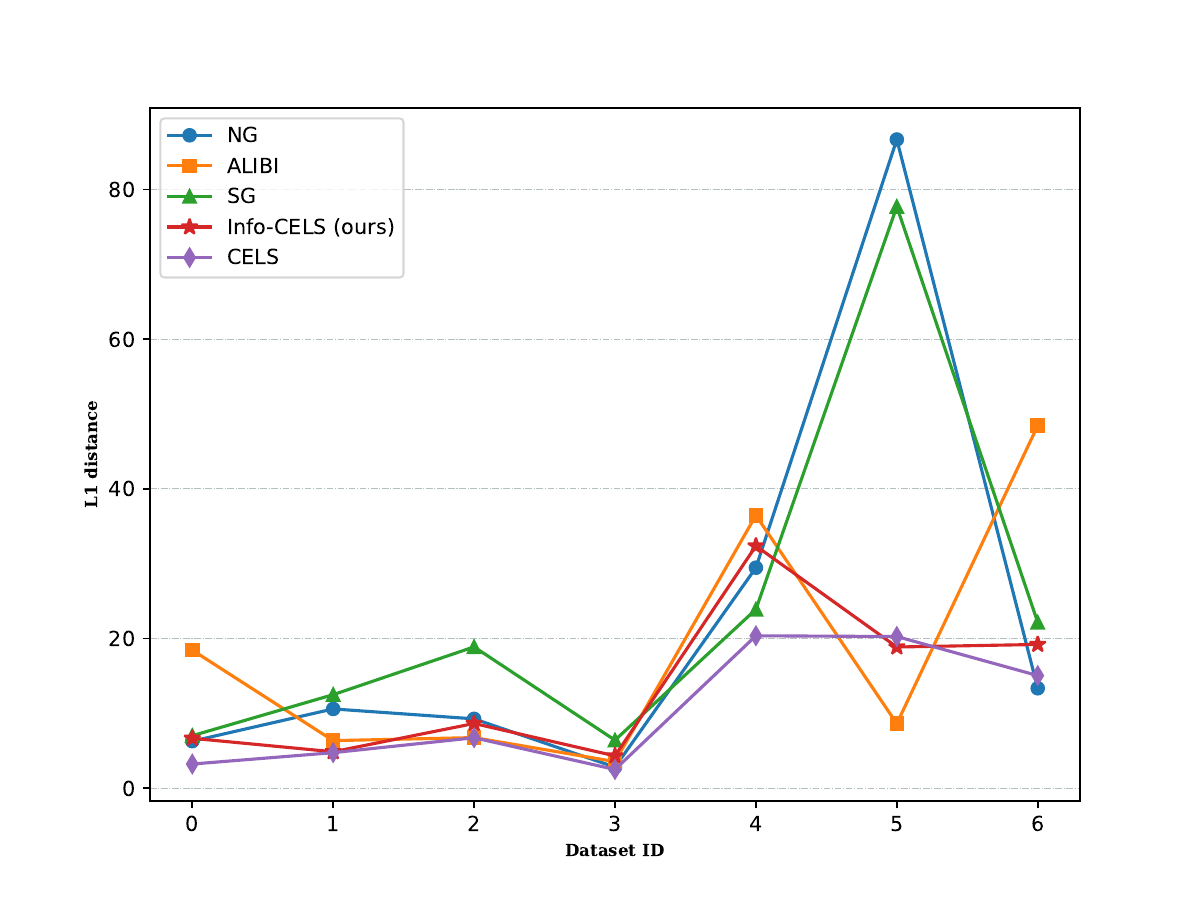}}% 2
\caption{Sparsity (the higher the better) and L1 distance (the lower the better) comparison of the CF explanations generated by ALIBI, NG, SG, CELS, and Info-CELS (All the reported results are the average value over the counterfactual set)}
    \label{fig: Spar_proximity}
\end{figure*}
% \vspace{-25pt}
% \vspace{-25pt}
\subsubsection{Sparsity and proximity evaluation: }
Figure \ref{fig: Spar_proximity} shows our experimental results in terms of sparsity and proximity across different explanation models. 

In Figure \ref{fig: Sparsity}, we observe the comparison of sparsity property among the generated counterfactuals across various explanation models. The sparsity metric represents the percentage of data points that remain unchanged after perturbation. Our proposed Info-CELS model demonstrates significantly higher sparsity levels compared to ALIBI and achieves sparsity levels comparable to SG. Although Info-CELS demonstrates lower sparsity compared to CELS and NG, it is noteworthy that CELS and NG achieve lower validity, indicating that some counterfactuals may not be valid due to insufficient perturbation. This suggests that while Info-CELS may compromise slightly on sparsity, it compensates by ensuring the validity of the generated counterfactual explanations, which is crucial for their usefulness and reliability in real-world applications. 

Figure \ref{fig: L1} evaluates the proximity property using the L1 distance metric. From this figure, we observe that Info-CELS achieves comparable proximity to the other baselines. Notably, Info-CELS stands out as the only model capable of generating counterfactuals with 100\% validity. This is a significant advantage, considering that other baselines may generate counterfactual instances with lower L1 distances but fail to ensure their validity.

Overall, the results suggest that Info-CELS strikes a balance between sparsity and proximity, achieving competitive performance while ensuring the validity of the generated counterfactual explanations. This highlights the effectiveness and reliability of Info-CELS in providing meaningful and trustworthy explanations for time series classification.

\subsection{Exploring interpretability\protect\footnote{Also referred to as plausibility (e.g., \cite{delaney2021instance}.)}}
In Figure \ref{fig:cfexample}, we compare the perturbations obtained when applying CELS and Info-CELS. As mentioned, Info-CELS produces counterfactuals with smoother transitions and fewer abrupt changes, indicating reduced noise and enhanced interpretability. Building on this observation, we further explore the plausibility of generated counterfactual explanations in time series data by employing novelty detection algorithms to detect out-of-distribution (OOD) explanations. Expanding on the concept of interpretability as we mentioned in section \ref{sec: pro}, we consider a counterfactual instance $\mathbf{x}'$ interpretable if it closely resembles the distribution of the model's training data. Specifically, $\mathbf{x}'$ should be classified as an inlier with respect to the training dataset. By leveraging novelty detection algorithms to assess the plausibility of counterfactual explanations, we aim to ensure that the generated explanations are not only valid but also realistic and trustworthy representations of potential alternative scenarios. Therefore, in this section, we implement the Isolation Forest (IF) \cite{liu2008isolation}, Local Outlier Factor Method (LOF) \cite{breunig2000lof}, and One-Class Support Vector Machine (OC-SVM) \cite{scholkopf2001estimating} to compare the plausibility of counterfactual explanations generated by CELS and Info-CELS. 

\begin{enumerate}
    \item \textbf{IF} is an algorithm used for anomaly detection, which works on the principle of isolating observations. Unlike many other techniques that rely on a distance or density measure, Isolation Forest is based on the idea that anomalies are 'few and different.'
    \item \textbf{LOF} is an algorithm used to identify density-based local outliers in a dataset. It works by comparing a point's local density with its neighbors' local densities to determine how isolated the point is.
    \item  \textbf{OC-SVM} is a variant of the Support Vector Machine (SVM) used for anomaly detection. It is an unsupervised learning algorithm designed to identify data points that are significantly different from the majority of the data.
\end{enumerate}

More experimental results about the other three baselines are shown on our project website \footnote{\href{https://sites.google.com/view/infocelsts/home}{https://sites.google.com/view/infocelsts/home}}.

Table \ref{tab: OOD} presents the percentage of generated counterfactuals that are identified as out-of-distribution instances. The results indicate that the percentage of out-of-distribution counterfactuals from Info-CELS is consistently lower than that from the CELS model. This finding suggests that counterfactual explanations produced by Info-CELS are more plausible and align better with the underlying data distribution compared to those generated by CELS. 

\begin{table}[htbp]
\centering
\caption{Comparing the CELS and info-CELS models on plausibility using three OOD metrics (IF, LOF, OC-SVM). Results indicate the percentage of generated counterfactuals that are out-of-distribution. (lower scores are better and the best are highlighted in bold).}
\label{tab: OOD}
\begin{tabular}{|l|ll|ll|ll|}
\hline
                 & \multicolumn{2}{c|}{\textbf{IF}}                   & \multicolumn{2}{c|}{\textbf{LOF}}                   & \multicolumn{2}{c|}{\textbf{OC-SVM}}                \\ \hline
\textbf{Dataset} & \multicolumn{1}{l|}{CELS}         & Info-CELS      & \multicolumn{1}{l|}{CELS}          & Info-CELS      & \multicolumn{1}{l|}{CELS}          & Info-CELS      \\ \hline
Coffee           & \multicolumn{1}{l|}{0.34}         & \textbf{0.318} & \multicolumn{1}{l|}{0.036}         & \textbf{0.036} & \multicolumn{1}{l|}{0.143}         & \textbf{0.107} \\ \hline
GunPoint         & \multicolumn{1}{l|}{0.214}        & \textbf{0.175} & \multicolumn{1}{l|}{0.267}         & \textbf{0.22}  & \multicolumn{1}{l|}{0.073}         & \textbf{0.047} \\ \hline
ECG200           & \multicolumn{1}{l|}{0.282}        & \textbf{0.220} & \multicolumn{1}{l|}{\textbf{0.02}} & \textbf{0.02}  & \multicolumn{1}{l|}{0.18}          & \textbf{0.13}  \\ \hline
TwoLeadECG       & \multicolumn{1}{l|}{0.381}        & \textbf{0.362} & \multicolumn{1}{l|}{0.02}          & \textbf{0}     & \multicolumn{1}{l|}{\textbf{0.22}} & \textbf{0.22}  \\ \hline
CBF              & \multicolumn{1}{l|}{0.159}        & \textbf{0.074} & \multicolumn{1}{l|}{\textbf{0}}    & \textbf{0}     & \multicolumn{1}{l|}{0.75}          & \textbf{0.38}  \\ \hline
BirdChicken & \multicolumn{1}{l|}{\textbf{0.5}} & \textbf{0.5}   & \multicolumn{1}{l|}{\textbf{0.05}} & \textbf{0.05}  & \multicolumn{1}{l|}{0.5}        & \textbf{0.3}  \\ \hline
Plane & \multicolumn{1}{l|}{0.33}         & \textbf{0.17}  & \multicolumn{1}{l|}{0.4}           & \textbf{0.31}  & \multicolumn{1}{l|}{0.35}          & \textbf{0.02}  \\ \hline
\end{tabular}
\end{table}

\subsection{Parameter Analysis}
\begin{figure*}[h!]
    \centering
    \begin{minipage}[b]{0.44\linewidth}  % Adjust the width as needed
        \centering
        \includegraphics[width=\linewidth]{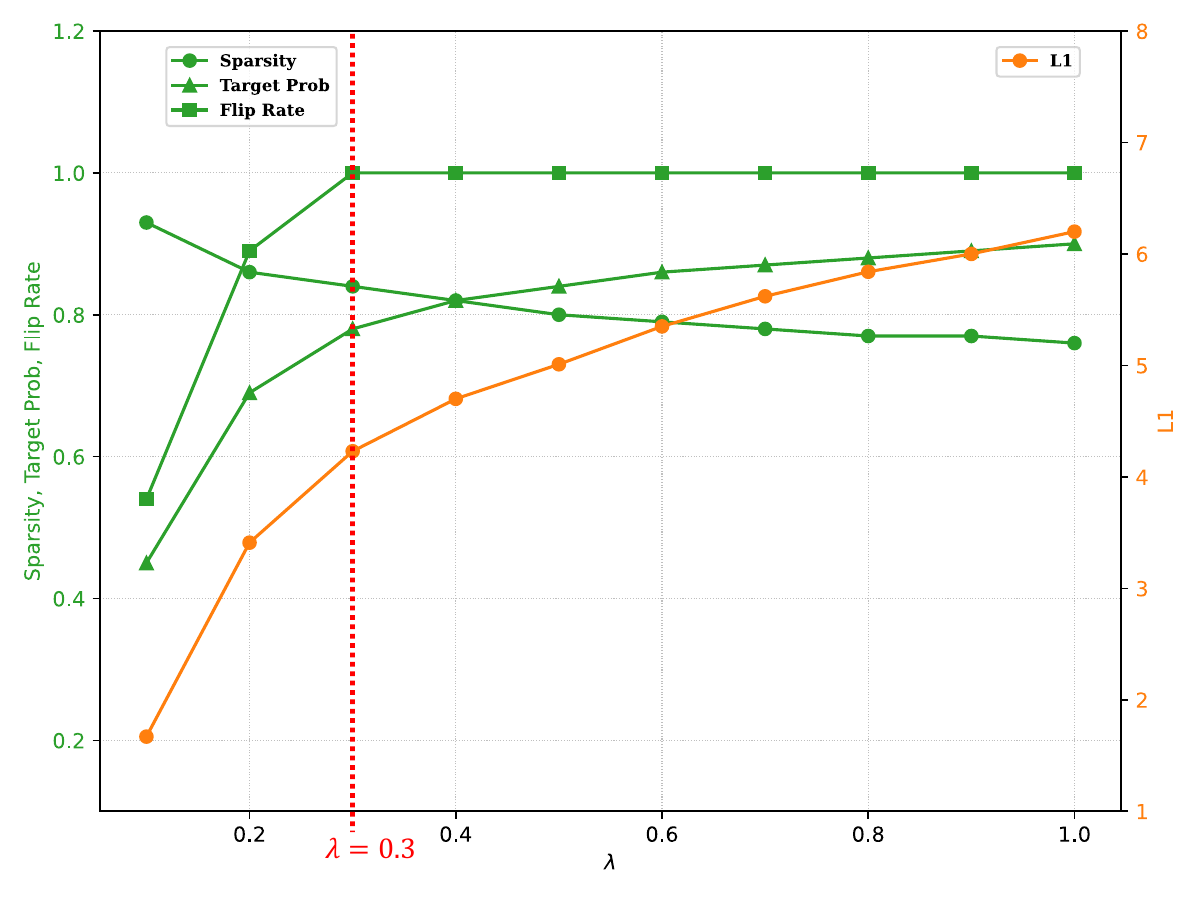}  % Adjust the file path as needed
        \caption{Different $\lambda$ values}
        \label{fig: lamb1}
    \end{minipage}
    \hfill
    \begin{minipage}[b]{0.49\linewidth}  % Adjust the width as needed
        \centering
        \includegraphics[width=\linewidth]{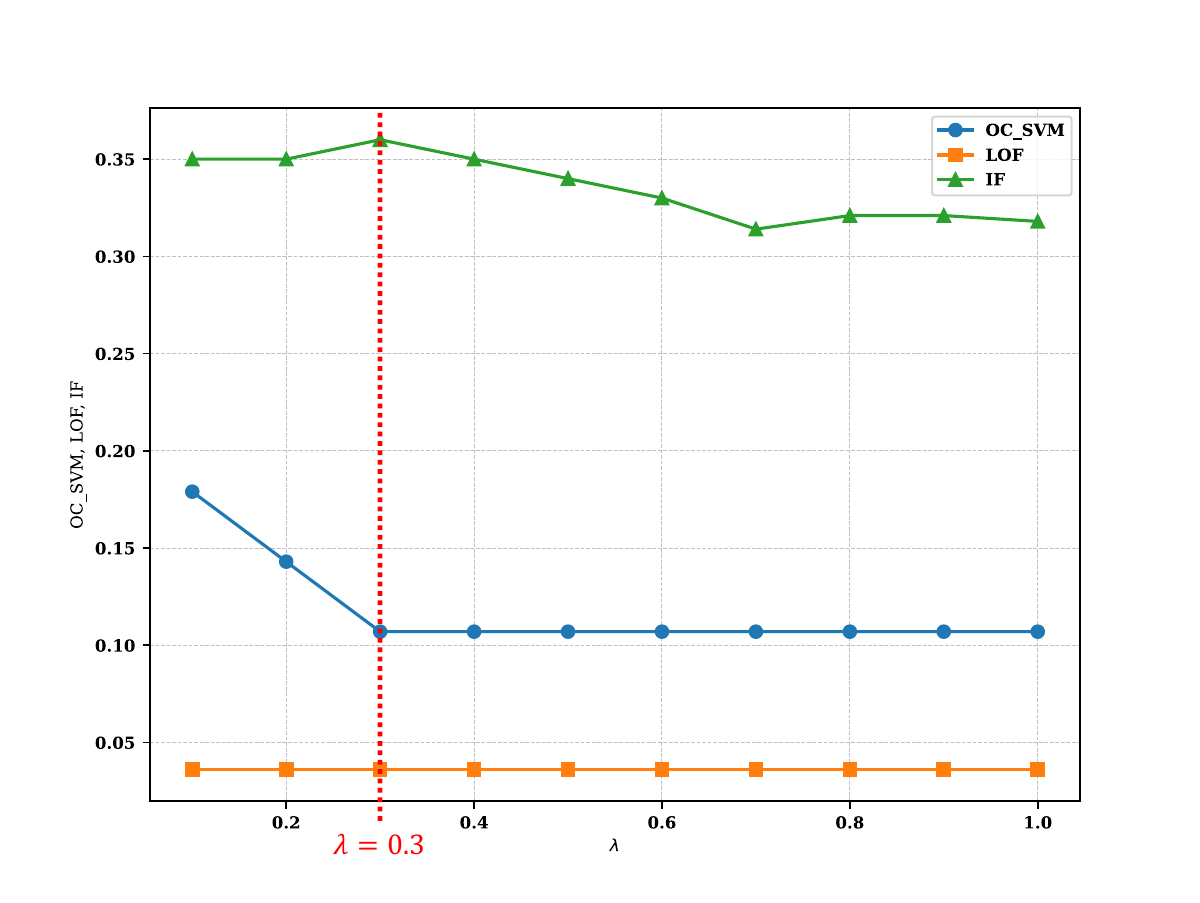}  % Adjust the file path as needed
        \caption{Different $\lambda$ values}
        \label{fig: lamb2}
    \end{minipage}
    \caption{Parameter analysis on the Coffee dataset.}
    \label{fig: para}
\end{figure*}

Note that we use $\lambda = 1$ in the previous experimental study section to ensure a high confidence level to get valid counterfactual explanations. In this section, we adopt the Coffee dataset to analyze the sensitivity of different $\lambda$ on the performance of validity, sparsity, proximity, and plausibility. In specific, we apply $\lambda = \{0.1, 0.2, 0.3, 0.4, 0.5, 0.6, 0.7, 0.8, 0.9, 1.0\}$ to Info-CELS to get the different counterfactuals on different $\lambda$ value settings. We then calculate the flip label rate, target probability, sparsity, L1 distance, and the three OOD metrics(IF, LOF, OC-SVM) under the corresponding $\lambda$ value. Figure \ref{fig: para} shows the results of applying different $\lambda$ on Info-CELs. 

From Figure \ref{fig: lamb1}, we can see that when $\lambda = 0.3$, we can already achieve 100\% validity with a decent target probability of about 0.8. When we increase the $\lambda$ value, the target probability tends to increase continuously. Referring to the equation provided (\ref{eq: max}), $L_{\text{Max}}$ is directly proportional to $\lambda$, meaning that as $\lambda$ increases, the impact of $L_{\text{Max}}$ on the overall loss becomes more pronounced. Since $L_{\text{Max}}$ is designed to maximize the class probability of the predicted counterfactual class, a higher weight assigned to this loss component results in a stronger emphasis on achieving a higher target probability. Therefore, as $\lambda$ increases, the model prioritizes maximizing the class probability, leading to a steady increase in the target probability. On the contrary, the L1 distance also keeps increasing when $\lambda$ is increasing, which results in lower proximity. This trend occurs because the $L_{\text{Budget}}$ loss component in the total loss function (\ref{eq: total loss}) becomes less influential with higher $\lambda$ values. As $\lambda$ increases, the weight assigned to $L_{\text{Budget}}$ becomes smaller relative to  $L_{\text{Max}}$ loss component. Consequently, the optimization process becomes less focused on minimizing the absolute values of $\boldsymbol{\theta}$ to reduce the overall magnitude of saliency changes, which leads to a larger L1 distance. Thus, as $\lambda$ increases, the L1 distance tends to increase accordingly. Similarly, the decrease in sparsity can also be attributed to the same reason.

From Figure \ref{fig: lamb2}, we can see that LOF is stable even though the $\lambda$ is changing and the OC-SVM also keeps stable when $\lambda>0.3$. The overall trend of IF is decreasing while the $\lambda$ is increasing, IF achieves the lowest value when $\lambda= 0.7$, and keeps stable with a slight increase after $\lambda>0.7$. 
% \vspace{-25pt}
% \vspace{-35pt}

\section{Conclusion}
In this paper, we introduce Info-CELS, the enhanced approach that builds upon CELS to solve the limitation of validity issues in the task of counterfactual explanation for time series classification. We extend beyond CELS to provide 100\% valid counterfactual explanations for time series classification. By removing the mask normalization step, Info-CELS effectively produces counterfactual explanations that exhibit 100\% validity with high sparsity, and proximity. In our experiments, we compare Info-CELS with CELS and three other state-of-the-art counterfactual explanation baseline models for time series classification tasks on seven popular datasets. Our experimental findings demonstrate that Info-CELS consistently yields significantly valid and highly interpretable counterfactual explanations. In the future, we want to explore the possibility of improving the contiguity property while maintaining high validity and interoperability.

\bibliographystyle{splncs04}
\bibliography{mybib}

\begin{thebibliography}{10}
\providecommand{\url}[1]{\texttt{#1}}
\providecommand{\urlprefix}{URL }
\providecommand{\doi}[1]{https://doi.org/#1}

\bibitem{adadi2018peeking}
Adadi, A., Berrada, M.: Peeking inside the black-box: a survey on explainable artificial intelligence (xai). IEEE access  \textbf{6},  52138--52160 (2018)

\bibitem{bahri2022shapelet}
Bahri, O., Boubrahimi, S.F., Hamdi, S.M.: Shapelet-based counterfactual explanations for multivariate time series. arXiv preprint arXiv:2208.10462  (2022)

\bibitem{bahri2022temporal}
Bahri, O., Li, P., Boubrahimi, S.F., Hamdi, S.M.: Temporal rule-based counterfactual explanations for multivariate time series. In: 2022 21st IEEE International Conference on Machine Learning and Applications (ICMLA). pp. 1244--1249. IEEE (2022)

\bibitem{breunig2000lof}
Breunig, M.M., Kriegel, H.P., Ng, R.T., Sander, J.: Lof: identifying density-based local outliers. In: Proceedings of the 2000 ACM SIGMOD international conference on Management of data. pp. 93--104 (2000)

\bibitem{chaovalitwongse2006electroencephalogram}
Chaovalitwongse, W.A., Prokopyev, O.A., Pardalos, P.M.: Electroencephalogram (eeg) time series classification: Applications in epilepsy. Annals of Operations Research  \textbf{148}(1),  227--250 (2006)

\bibitem{dau2019ucr}
Dau, H.A., Bagnall, A., Kamgar, K., Yeh, C.C.M., Zhu, Y., Gharghabi, S., Ratanamahatana, C.A., Keogh, E.: The ucr time series archive. IEEE/CAA Journal of Automatica Sinica  \textbf{6}(6),  1293--1305 (2019)

\bibitem{delaney2021instance}
Delaney, E., Greene, D., Keane, M.T.: Instance-based counterfactual explanations for time series classification. In: International Conference on Case-Based Reasoning. pp. 32--47. Springer (2021)

\bibitem{filali2022mining}
Filali~Boubrahimi, S., Hamdi, S.M.: On the mining of time series data counterfactual explanations using barycenters. In: Proceedings of the 31st ACM International Conference on Information \& Knowledge Management. pp. 3943--3947 (2022)

\bibitem{guidotti2018local}
Guidotti, R., Monreale, A., Ruggieri, S., Pedreschi, D., Turini, F., Giannotti, F.: Local rule-based explanations of black box decision systems. arXiv preprint arXiv:1805.10820  (2018)

\bibitem{hosseinzadeh2023metforc}
Hosseinzadeh, P., Bahri, O., Li, P., Boubrahimi, S.F., Hamdi, S.M.: Metforc: Classification with meta-learning and multimodal stratified time series forest. In: 2023 International Conference on Machine Learning and Applications (ICMLA). pp. 1248--1252. IEEE (2023)

\bibitem{ismail2019deep}
Ismail~Fawaz, H., Forestier, G., Weber, J., Idoumghar, L., Muller, P.A.: Deep learning for time series classification: a review. Data mining and knowledge discovery  \textbf{33}(4),  917--963 (2019)

\bibitem{kingma2014adam}
Kingma, D.P., Ba, J.: Adam: A method for stochastic optimization. arXiv preprint arXiv:1412.6980  (2014)

\bibitem{li2022fast}
Li, P., Bahri, O., Boubrahimi, S.F., Hamdi, S.M.: Fast counterfactual explanation for solar flare prediction. In: 2022 21st IEEE International Conference on Machine Learning and Applications (ICMLA). pp. 1238--1243. IEEE (2022)

\bibitem{li2022sg}
Li, P., Bahri, O., Boubrahimi, S.F., Hamdi, S.M.: Sg-cf: Shapelet-guided counterfactual explanation for time series classification. In: 2022 IEEE International Conference on Big Data (Big Data). pp. 1564--1569. IEEE (2022)

\bibitem{li2023cels}
Li, P., Bahri, O., Boubrahimi, S.F., Hamdi, S.M.: Cels: Counterfactual explanations for time series data via learned saliency maps. In: 2023 IEEE International Conference on Big Data (BigData). pp. 718--727. IEEE (2023)

\bibitem{li2022motif}
Li, P., Boubrahimi, S.F., Hamd, S.M.: Motif-guided time series counterfactual explanations. arXiv preprint arXiv:2211.04411  (2022)

\bibitem{liu2008isolation}
Liu, F.T., Ting, K.M., Zhou, Z.H.: Isolation forest. In: 2008 eighth ieee international conference on data mining. pp. 413--422. IEEE (2008)

\bibitem{lundberg2017unified}
Lundberg, S.M., Lee, S.I.: A unified approach to interpreting model predictions. Advances in neural information processing systems  \textbf{30} (2017)

\bibitem{mahendran2015understanding}
Mahendran, A., Vedaldi, A.: Understanding deep image representations by inverting them. In: Proceedings of the IEEE conference on computer vision and pattern recognition. pp. 5188--5196 (2015)

\bibitem{mothilal2020explaining}
Mothilal, R.K., Sharma, A., Tan, C.: Explaining machine learning classifiers through diverse counterfactual explanations. In: Proceedings of the 2020 conference on fairness, accountability, and transparency. pp. 607--617 (2020)

\bibitem{ribeiro2016should}
Ribeiro, M.T., Singh, S., Guestrin, C.: " why should i trust you?" explaining the predictions of any classifier. In: Proceedings of the 22nd ACM SIGKDD international conference on knowledge discovery and data mining. pp. 1135--1144 (2016)

\bibitem{scholkopf2001estimating}
Sch{\"o}lkopf, B., Platt, J.C., Shawe-Taylor, J., Smola, A.J., Williamson, R.C.: Estimating the support of a high-dimensional distribution. Neural computation  \textbf{13}(7),  1443--1471 (2001)

\bibitem{tjoa2020survey}
Tjoa, E., Guan, C.: A survey on explainable artificial intelligence (xai): Toward medical xai. IEEE transactions on neural networks and learning systems  \textbf{32}(11),  4793--4813 (2020)

\bibitem{wachter2017counterfactual}
Wachter, S., Mittelstadt, B., Russell, C.: Counterfactual explanations without opening the black box: Automated decisions and the gdpr. Harv. JL \& Tech.  \textbf{31}, ~841 (2017)

\bibitem{wang2017time}
Wang, Z., Yan, W., Oates, T.: Time series classification from scratch with deep neural networks: A strong baseline. In: 2017 International joint conference on neural networks (IJCNN). pp. 1578--1585. IEEE (2017)

\end{thebibliography}


% LNCS Homepage, \url{http://www.springer.com/lncs}. Last accessed 4
\end{document}